\def\year{2021}\relax
\documentclass[letterpaper]{article} % DO NOT CHANGE THIS
\usepackage{aaai21}  % DO NOT CHANGE THIS
\usepackage{times}  % DO NOT CHANGE THIS
\usepackage{helvet} % DO NOT CHANGE THIS
\usepackage{courier}  % DO NOT CHANGE THIS
\usepackage[hyphens]{url}  % DO NOT CHANGE THIS
\usepackage{graphicx} % DO NOT CHANGE THIS
\def\UrlFont{\rm}  % DO NOT CHANGE THIS
\usepackage{natbib}  % DO NOT CHANGE THIS AND DO NOT ADD ANY OPTIONS TO IT
\usepackage{caption} % DO NOT CHANGE THIS AND DO NOT ADD ANY OPTIONS TO IT
\usepackage[switch]{lineno}  %

\usepackage[utf8]{inputenc} % allow utf-8 input
\usepackage{booktabs}       % professional-quality tables
\usepackage{amsfonts}       % blackboard math symbols
\usepackage{nicefrac}       % compact symbols for 1/2, etc.
\usepackage{microtype}      % microtypography
\usepackage{amssymb}

\usepackage{soul}
\usepackage{amsmath}
\usepackage{amsthm}
\usepackage{algorithm}
\usepackage{algorithmic}

\usepackage{subcaption}
\graphicspath{{fig/}}

\newtheorem{example}{Example}

 \newcounter{myenum}
\newenvironment{flushenumerate}{%
 \begin{list}{\arabic{myenum}.}%
   {\setlength{\leftmargin}{25pt}}%
    \setlength{\labelwidth}{20pt}
    \setlength{\itemindent}{7pt}
    \setlength{\labelsep}{0.5em}
 \setlength{\itemsep}{1pt}
 \setlength{\parskip}{0pt}
 \setlength{\parsep}{0pt}
    \usecounter{myenum}}%
 {\end{list}}
\usepackage{bbm}

\title{Encoding Human Domain Knowledge to Warm Start Reinforcement Learning}

\author{ Andrew Silva, Matthew Gombolay \\}
\affiliations{
  Institute for Robotics and Intelligent Machines \\
    Georgia Institute of Technology \\
    andrew.silva@gatech.edu, matthew.gombolay@cc.gatech.edu
}
\begin{document}
% \linenumbers
\maketitle

\begin{abstract}
Deep reinforcement learning has been successful in a variety of tasks, such as game playing and robotic manipulation. However, attempting to learn \textit{tabula rasa} disregards the logical structure of many domains as well as the wealth of readily available knowledge from domain experts that could help ``warm start'' the learning process.
%Learning-from-demonstration techniques are also inefficient in translating this domain knowledge.
We present a novel reinforcement learning technique that allows for intelligent initialization of a neural network weights and architecture. Our approach permits the encoding domain knowledge directly into a neural decision tree, and improves upon that knowledge with policy gradient updates. We empirically validate our approach on two OpenAI Gym tasks and two modified StarCraft 2 tasks, showing that our novel architecture outperforms multilayer-perceptron and recurrent architectures. Our knowledge-based framework finds superior policies compared to imitation learning-based and prior knowledge-based approaches. Importantly, we demonstrate that our approach can be used by untrained humans to initially provide $>80\%$ increase in expected reward relative to baselines prior to training ($p < 0.001$), which results in a $>60\%$ increase in expected reward after policy optimization ($p = 0.011$).
\end{abstract}

\section{Introduction}
\label{sec:intro}
%Reinforcement learning (RL) has seen impressive success in many domains~\cite{Andrychowicz2018LearningDI,espeholt2018impala,sun2018tstarbots}. 
As reinforcement learning (RL) is applied to increasingly complex domains, such as real-time strategy games
%~\cite{sun2018tstarbots,alphastarblog}
or robotic manipulation
%~\cite{Andrychowicz2018LearningDI,rajeswaran2017learning}, 
%current state-of-the-art approaches require incredible amounts of compute power, data, and compute time which are unavailable to most practitioners across academia and industry. Further,
RL and imitation learning (IL) approaches fail to quickly capture the wealth of expert knowledge that already exists for many domains. 
Existing approaches to using IL as a warm start require large datasets or tedious human labeling as the agent learns everything, from vision to control to policy, all at once.
Unfortunately, these large datasets often do not exist, as collecting these data is impractical or expensive, and humans will not patiently label data for IL-based agents \cite{amershi2014power}.
% This failure leads to state-of-the-art approaches which require enormous compute power, data, or time.
While humans may not label enough state-action pairs to train IL-based agents , there is an opportunity to improve warm starts by soliciting expertise from a human once, and then leveraging this expertise to initialize an RL agent's neural network architecture and policy. With this approach, we circumvent the need for IL and instead directly imbue human expertise into an RL agent.

%We aim to solve this shortcoming by integrating human knowledge into our learning algorithms. While human domain experts have a variety of strategies or heuristics that underlie much of their success, a mechanism to translate this knowledge directly into a learning framework is missing in current deep RL literature. While knowledge-based learning approaches exist \cite{humbird2018deep,towell1994knowledge,zhang2019leveraging}, none directly initialize an both an agent's architecture \textit{and} policy with domain knowledge.

To achieve this blending of human domain knowledge with the strengths of RL, we propose Propositional Logic Nets (\textsc{ProLoNets}), a new approach to directly encode domain knowledge as a set of propositional rules into a neural network, as depicted in Figure \ref{fig:study-ui}. 
Our approach leverages decision tree policies from humans to directly initialize a neural network (Figure \ref{fig:architecture}). 
We use decision trees to allow humans to specify behaviors to guide the agent through a given domain, such as high-level instructions for keeping a pole balanced on the cart pole problem. %\cite{barto1983neuronlike}.
Importantly, this policy specification does not require the human to demonstrate the balancing act in all possible states, nor does it require the human to label actions as being ``good'' or ``bad.'' 

By directly imbuing logical propositions from the tree into neural network weights, an RL agent can immediately begin learning productive strategies.
%rather than expending hundreds of CPU hours in a complex domain while lacking the ability to explore it in a meaningful way. 
This approach leverages readily available domain knowledge while still retaining the ability to learn and improve over time, eventually outperforming the expertise with which it was initialized. By exploiting the structural and logical rules inherent to many tasks to which RL is applied, we can bypass early random exploration and expedite an agent's learning in a new domain. 

We demonstrate that our approach can outperform standard deep RL across two OpenAI gym domains \cite{1606.01540} and two modified StarCraft II domains \cite{vinyals2017starcraft}, and that our framework is superior to state-of-the-art, IL-based RL, even with observation of that same domain expert knowledge. Finally, in a wildfire simulation domain, we show that our framework can work with untrained human participants. 
Our three primary contributions include:
% We make three primary contributions in this work.

% \nocite{crandall2006working}
% \nocite{newell1983psychology}
\begin{figure*}[t]
\centering
        \includegraphics[width=0.7\linewidth]{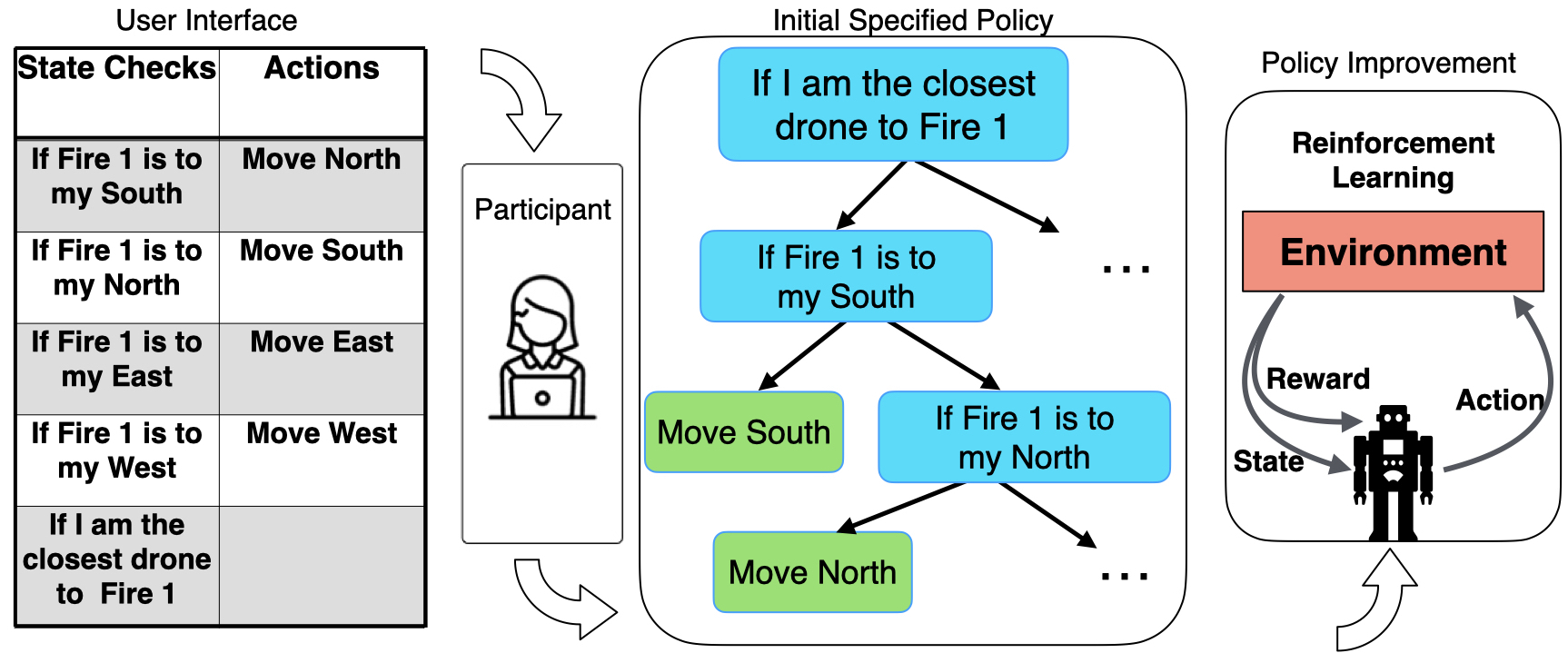}
        \caption{A visualization of our approach as it applies to our user study. Participants interact with a UI of state-checks and actions to construct a decision tree policy that is then used to directly initialize a \textsc{ProLoNet}'s architecture and parameters. The \textsc{ProLoNet} can then begin reinforcement learning in the given domain, outgrowing its original specification.}
        \label{fig:study-ui}
\end{figure*}

\begin{enumerate}
    \item We formulate a novel approach for capturing human domain expertise in a trainable RL framework via our architecture, \textsc{ProLoNets}, which we show outperforms baseline RL approaches, including IL-based \cite{cheng2018fast} and knowledge-based techniques \cite{humbird2018deep}, obtaining $>100\%$ more average reward on a StarCraft 2 mini-game.
    \item We introduce dynamic growth to \textsc{ProLoNets}, enabling greater expressivity over time to surpass original initializations and yielding twice as much average reward in the lunar lander domain.
    \item We conduct a user study in which non-expert humans leveraged \textsc{ProLoNets} to specify policies that resulted in higher cumulative rewards, both before and after training, relative to all baselines ($p < 0.05$).
\end{enumerate}
\section{Related work}
\label{sec:related-work}

Warm starts have been used for RL ~\cite{cheng2018fast,aaai2019lecture,zhu2017effective} as well as in supervised learning for many tasks \cite{garcez2012neural,hu2016harnessing,kontschieder2015deep,wang2017combining}. While these warm start or knowledge-based systems have provided an interesting insight into the efficacy of warm starts or human-in-the-loop learning in various domains, these systems typically involve either large labeled datasets with tedious human labeling and feedback, or they require some automated oracle to label actions as ``good'' or ``bad.'' In highly challenging domains or problems, building such an oracle is rarely feasible. Moreover, it is not always possible to acquire a large labeled dataset for new domains.
However, it is often possible to solicit a policy from a human in the form of a high-level series of if-then checks in critical states. These decisions can be collected as a decision tree. 
Our research seeks to convert decision tree into a neural network for RL.

Researchers have previously sought to bridge the gap between decision trees and deep networks~\cite{humbird2018deep,kontschieder2015deep,laptev2014convolutional}. This work has focused on either partitioning a subspace of the data for more efficient inference~\cite{tanno2018adaptive}, to enable more explicit interpretability by visualizing a network's classification policy~\cite{frosst2017distilling,silva2019optimization}, or for warm starting through supervised pre-training on labeled data. As discussed, this data may not be available thus creating a need for methods which can solicit this initialization tree directly from a human.

Most closely related to our work is deep jointly-informed neural networks (DJINN) \cite{humbird2018deep}, which is the latest in a long line of knowledge-based neural network research \cite{francca2014fast,garcez2012neural,maclin1996creating,richardson2006markov,towell1994knowledge}. DJINN uses a decision tree learned over a training set in order to initialize the structure of a network's hidden layers and to route input data appropriately. However, DJINN does not explicitly initialize rules, nor does it leverage rules solicited from humans.
This distinction means that DJINN creates an architecture for routing information appropriately, but the decision-criteria in each layer must be learned from scratch. % the routed information is effectively meaningless and
Our work, on the other hand, directly initializes both the structure \emph{and} the rules of a neural network, meaning that the human's expertise is more completely leveraged for a more useful warm start in RL domains. We build on decades of research demonstrating the value of human-in-the-loop learning \cite{towell1994knowledge,zhang2019leveraging} to leverage logical rules solicited from humans in the form of a decision tree to intelligently initialize the structure and rules of a deep network. 

Our work is related to IL and to knowledge-based or human-in-the-loop RL frameworks \cite{zhang2019leveraging, aaai2019lecture,macglashan2017interactive} and apprenticeship learning and IRL \cite{abbeel2004apprenticeship,knox2009interactively}. Importantly, however, our approach does not require demonstrations or datasets to mimic human behavior. While our approach directly initializes with a human-specified policy, IL methods require large labeled datasets \cite{edwards2018imitating} or an oracle to label data before transitioning to RL, as in the LOKI \cite{cheng2018fast} framework. 
%Yet even with a pre-trained policy, LOKI still requires extensive domain experience before beginning the RL stage. While a human can act as an oracle for IL, it is unrealistic for a human to patiently label replay data for any great length of time~\cite{amershi2014power}. 
Our approach translates human expertise directly into an RL agent's policy and begins learning immediately, sidestepping the IL and labeling phase.

\section{Preliminaries}
\label{sec:preliminaries}

Within RL, we consider problems presented as a Markov decision process (MDP), which is a 5-tuple $\langle S, A, T, R, \lambda \rangle$ where $s \in S$ are states drawn from the state space or domain, $a \in A$ are possible actions drawn from the action space, $T(s', a, s)$ is the transition function representing the likelihood of reaching a next state $s'$ by taking some action $a$ in a given state $s$, $R(s)$ is the reward function which determines the reward for each state, and $\lambda$ is a discount factor. In this work, we examine discrete action spaces and semantically meaningful state spaces-- intelligent initialization for continuous outputs and unstructured inputs is left to future work. The goal of our RL agent is to find a policy, $\pi(a|s)$, that selects actions in states to maximize the agent's expected long-term cumulative reward. IL approaches, such as ILPO~\cite{edwards2018imitating}, operate under a similar framework, though they do not make use of the reward signal and instead perform supervised learning according to oracle data. 

\section{Approach}
\label{sec:approach} 

\begin{figure}[t]
    \centering
         \includegraphics[width=\linewidth]{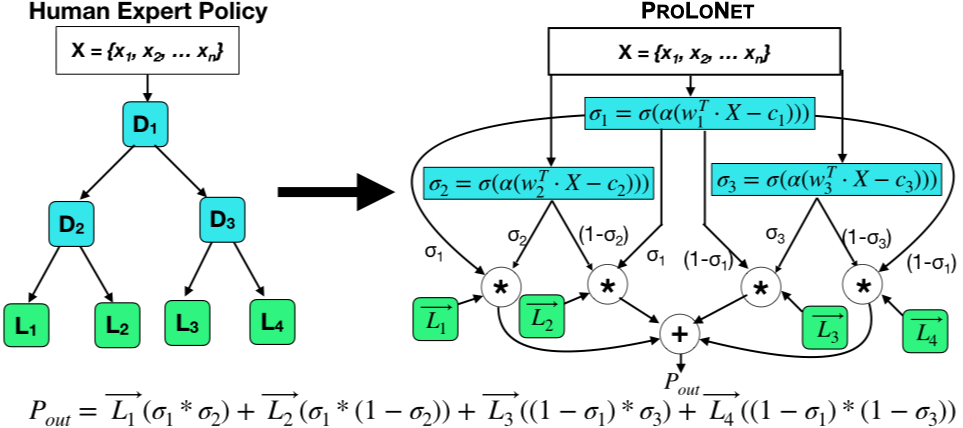}
    \caption{A traditional decision tree and a \textsc{ProLoNet}. Decision nodes become linear layers, leaves become action weights, and the final output is a sum of the leaves weighted by path probabilities.}
    \label{fig:architecture}
\end{figure}

We provide a visual overview of the \textsc{ProLoNet} architecture in Figure \ref{fig:architecture}. To intelligently initialize a \textsc{ProLoNet}, a human user first provides a policy in the form of some hierarchical set of decisions. These policies are solicited through simple user interactions for specifying instructions, as in Section \ref{sec:user-study}. The user's decision-making process is then translated into a set of weights $\vec{w_n} \in W$ and comparator values $c_n \in C$ representing each rule, shown in Algorithm \ref{alg:prolo-init}. Each weight $\vec{w_n}$ determines which input features to consider, and, optionally, how to weight them, as there is a unique weight value for each input feature (i.e. $|\vec{w_n}|==|S|$ for an input space $S$). The comparator $c_n$ is used as a threshold for the weighted features. 

Each decision node $D_n$ throughout the network is represented as $D_{n} = \sigma [\alpha(\vec{w_n}^T * \vec{X} - c_n)]$, where $\vec{X}$ is the input data, $\sigma$ is the sigmoid function, and $\alpha$ serves to throttle the confidence of decision nodes. Less confidence in the tree allows for more uncertainty in decision making~\cite{yuan1995induction}, leading to more exploration, even from an expert initialization. High values of $\alpha$ emphasize the difference between the comparator and the weighted input, thus pushing the tree to be more boolean. Lower values of $\alpha$ encourage a smoother tree, with $\alpha=0$ producing uniformly random decisions. We allow $\alpha$ to be a learned parameter.

\begin{example}[\textsc{ProLoNet} Initialization]
\label{ex:init}
Assume we are in the cart pole domain \cite{barto1983neuronlike} and have solicited the following from a human: ``If the cart's $x$ position is right of center, move left; otherwise, move right,'' and that the user indicates $x\_position$ is the first input feature of four and that the center is at 0. We therefore initialize our primary node $D_0$ with $\vec{w_0} = [1, 0, 0, 0]$ and $c_0 = 0$, following lines 5-8 in Alg. \ref{alg:prolo-init}. Following lines 11-13, we create a new leaf $\vec{l_0} = [1, 0]$ (Move Left) and a new leaf $\vec{l_1} = [0, 1]$ (Move Right). Finally, we set the paths $Z(\vec{l_0}) = D_0$ and  $Z(\vec{l_1}) = (\neg \, D_0)$. The resulting probability distribution over the agent's actions is a softmax over $(D_0*\vec{l_0} + (1-D_0)*\vec{l_1})$. 
\end{example}

\begin{algorithm}[H]
  \caption{Intelligent Initialization}
  \label{alg:prolo-init}

\begin{algorithmic}[1]
  \STATE {\bfseries Input:} Expert Propositional Rules $R_d$
  \STATE {\bfseries Input:} Input Size $I_S$, Output Size $O_S$
  \STATE $W, C, L =$ \{\}
  \FOR{$r \in R_d$}
  \IF{$r$ is a state check}
  \STATE $\mathbf{s} =$ feature index in $r$
  \STATE $w = \vec{0}^{I_S}$, $w[\mathbf{s}] = 1$
  \STATE $c = $ comparison value in $r$
  \STATE $W = W \cup w$, $C = C \cup c$
  \ENDIF
  \IF{$r$ is an action}
  \STATE $\mathbf{a} =$ action index in $r$
  \STATE $l = \vec{0}^{O_S}$, $l[\mathbf{a}] = 1$
  \STATE $L = L \cup l$
  \ENDIF
  \ENDFOR
  \STATE {\bfseries Return:} $W$, $C$, $L$
\end{algorithmic}
\end{algorithm}

\begin{algorithm}[H]
  \caption{Dynamic Growth}
  \label{alg:deepeninng}

\begin{algorithmic}[1]
  \STATE {\bfseries Input:} \textsc{ProLoNet} $P_d$
  \STATE {\bfseries Input:} Deeper \textsc{ProLoNet} $P_{d+1}$
  \STATE {\bfseries Input:} $\epsilon =$ minimum confidence 
  \STATE $H(\vec{l_i}) = $ Entropy of leaf $\vec{l_i}$,
  \FOR{$l_i \in L \in P_d$}
  \STATE Calculate $H(l_i)$
  \STATE Calculate $H(l_{d1})$, $H(l_{d2})$ \\ for leaves under $l_i$ in $P_{d+1}$
  \IF{$H(l_i) > (H(l_{d1}) + H(l_{d2}) + \epsilon)$}
  \STATE Deepen $P_d$ at $l_i$ using $l_{d1}$ and $l_{d2}$
  \STATE Deepen $P_{d+1}$ at $l_{d1}$ and $l_{d2}$ randomly
  \ENDIF
  \ENDFOR
\end{algorithmic}
\end{algorithm}

After all decision nodes are processed, the values of $D_n$ from each node represent the likelihood of that condition being $TRUE$. In contrast, $(1-D_n)$ represents the likelihood of the condition being $FALSE$. With these likelihoods, the network then multiplies out the probabilities for different paths to all leaf nodes. Every leaf $\vec{l} \in L$ contains a path $z \in Z$, a set of decision nodes which should be $TRUE$ or $FALSE$ in order to reach $\vec{l}$, as well as a prior set of weights for each output action $a \in \vec{a}$. For example, in Figure \ref{fig:architecture}, $z_1 = D_1 * D_2$, and $z_3 = (1-D_1)*D_3$. The likelihood of each action $a$ in leaf $\vec{l_i}$ is determined by multiplying the probability of reaching leaf $\vec{l_i}$ by the prior weight of the outputs within leaf $\vec{l_i}$. After calculating the outputs for every leaf, the leaves are summed and passed through a softmax function to provide the final output distribution.

\begin{example}[\textsc{ProLoNet} Inference]
\label{ex:inference}

Consider an example cart pole state, X=[2, 1, 0, 3] passed to the \textsc{ProLoNet} from Example \ref{ex:init}. Following $D_{n} = \sigma [\alpha(\vec{w_n}^T * \vec{X} - c_n)]$, the network arrives at $\sigma([1,0,0,0]*[2,1,0,3]-0)=0.88$ for $D_0$, meaning “mostly true.” This probability propagates to the two leaf nodes using their respective paths, making the output of the network a probability given by $(0.88 * [1, 0] + (1-0.88)*[0, 1]) = [0.88, 0.12]$. Accordingly, the agent selects the first action with probability 0.88 and the second action otherwise. An algorithmic expression of the forward-pass is provided in the supplementary material.
\end{example}

\paragraph{Dynamic Growth --}
\label{sub:deepening}

\begin{figure}
    \centering
    \includegraphics[width=\linewidth]{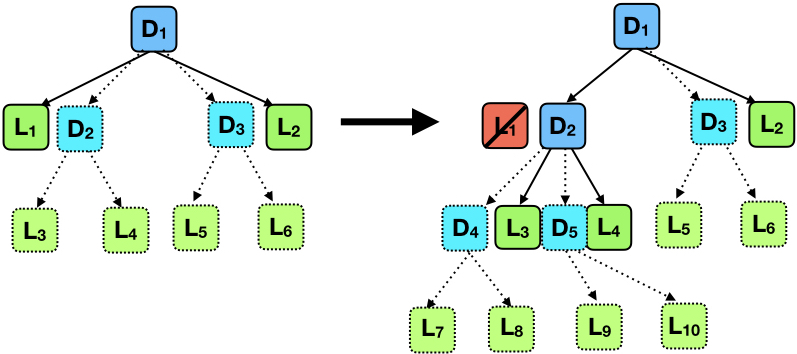}
    \caption{The dynamic growth process with a deeper \textsc{ProLoNet} shown in paler colors and dashed lines. When $H(L_3)+H(L_4) < H(L_1)$, the agent replaces $L_1$ with $D_2$, $L_3$, and $L_4$ and adds a new level to the deeper actor.} % and $L_4$ combined have lower entropy that $L_1$
    \label{fig:example-deepen}
\end{figure}

\textsc{ProLoNets} are able to follow expert strategies immediately, but they may lack the expressive capacity to learn more optimal policies once they are deployed into a domain. If an expert policy involves a small number of decisions, the network will have a small number of weight vectors and comparators to use for its entire existence. To enable the \textsc{ProLoNet} architecture to continue to grow beyond its initial definition, we introduce a dynamic growth procedure, which is outlined in Algorithm \ref{alg:deepeninng} and Figure \ref{fig:example-deepen}. 

Upon initialization, a \textsc{ProLoNet} agent maintains two copies of its actor. The first is the shallower, unaltered initialized version, and the second is a deeper version in which each leaf is transformed into a randomly initialized decision node with two new randomly initialized leaves (line 1 of Alg. \ref{alg:deepeninng}). This deeper agent has more parameters to potentially learn more complex policies, but at the cost of added randomness and uncertainty, reducing the utility of the intelligent initialization.

As the agent interacts with its environment, it relies on the shallower network to generate actions, as the shallow network represents the human's domain knowledge. After each episode, the off-policy update is run over the shallower and deeper networks. Finally, after the off-policy updates, the agent compares the entropy of the shallower actor's leaves to the entropy of the deeper actor's leaves and selectively deepens when the leaves of the deeper actor are less uniform than those of the shallower actor (lines 3-7). We find that this dynamic growth mechanism improves stability and average cumulative reward.

\begin{example}[\textsc{ProLoNet} Dynamic Growth]
\label{ex:deepen}

Assume the cart pole agent’s shallower actor has found a local minimum with $l_1=[0.5, 0.5]$, while the deeper actor has $l_3=[0.9, 0.1]$ and $l_4=[0.1, 0.9]$. Seeing that $l_1$ is offering little benefit to the current policy, and $D_2$ in the deeper actor is able to make a decision about which action offers the most reward, the agent would dynamically deepen at $l_1$, copying over the deeper actor's parameters and becoming more decisive in that area of its policy. The deeper actor would also grow with a random set of new parameters, as shown in Figure \ref{fig:example-deepen}.
\end{example}

\section{Experimental evaluation}
\label{sec:experiments}

We conduct two complementary evaluations of the \textsc{ProLoNet} as a framework for RL with human initialization. The first is a controlled investigation with expert initialization in which an author designs heuristics for a set of domains with varying complexity; this allows us to confirm that our architecture is competitive with baseline learning frameworks. We also perform an ablation of intelligent initialization and dynamic growth in this set of experiments.  The second evaluation is a user study to support our claim that untrained users can specify policies that serve to improve RL. 

In our first evaluation, we assess our algorithm in StarCraft II (SC2) for macro and micro battles as well as the OpenAI Gym \cite{1606.01540} lunar lander and cart pole environments. Optimization details, hyperparameters, and code are all provided in the supplementary material. 

%All actors are updated with proximal policy optimization (PPO)~\cite{schulman2017proximal}. Notably, for the two SC2 domains, we find that multiplying the PPO update by the Kullback-Leibler divergence between old and new policies yields superior performance. The critic's loss function is the mean-squared error between the output of the critic and the reward from the state-action pair. All approaches are trained with RMSProp~\cite{tieleman2012lecture}. We set our reward discount factor to 0.99, learning rates to 1e-2 for Gym environments, and 1e-4 for the SC2 domains, following a hyperparameter search between 1-e2 and 1e-5. Update batch sizes dynamically grow as more replay experience is available. In all domains, the \textsc{ProLoNet} $\alpha$ parameter is initialized to 1.  Our agents utilize two separate networks: one for the actor and one for the critic. For our approach, the critic network is initialized as a copy of the actor as we do not solicit intelligent value predictions, only policies. Our dynamic growth hyperparameter $\epsilon$ is set to $\epsilon=0.1$ based upon experimental observation.

To evaluate the impact of dynamic growth and intelligent initialization, we perform an ablation study and include results from these experiments in Table \ref{tab:ablate}. For each $N$-mistake agent, weights, comparators, and leaves are randomly negated according to $N$, up to a maximum of $2N$ for each category.
%For example, ``N = 0.05'' has a maximum of 10\% of its weights, comparators, and leaves negated, each with probability $0.05$.

\begin{figure*}[t]
\centering
\begin{subfigure}[b]{\textwidth}
    \centering
    \includegraphics[width=0.6\textwidth]{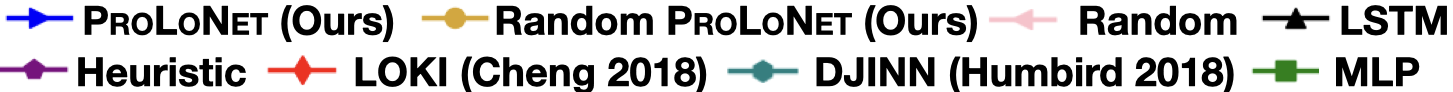}
\end{subfigure} 
    \begin{subfigure}[b]{0.31\textwidth}
        \includegraphics[width=\textwidth]{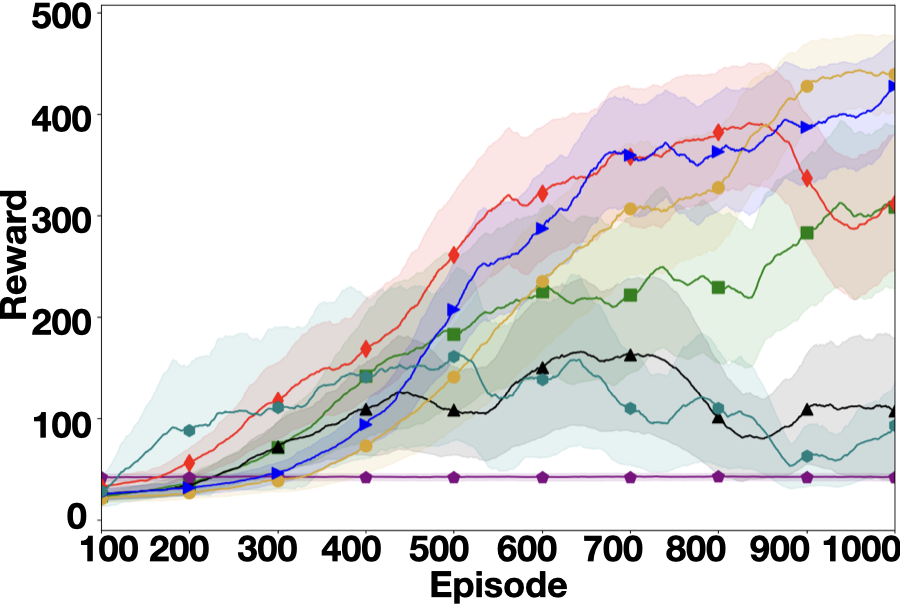}
        \caption{Cart Pole}
        \label{fig:cart-results}
    \end{subfigure}
    ~~
    \begin{subfigure}[b]{0.31\textwidth}
        \includegraphics[width=\textwidth]{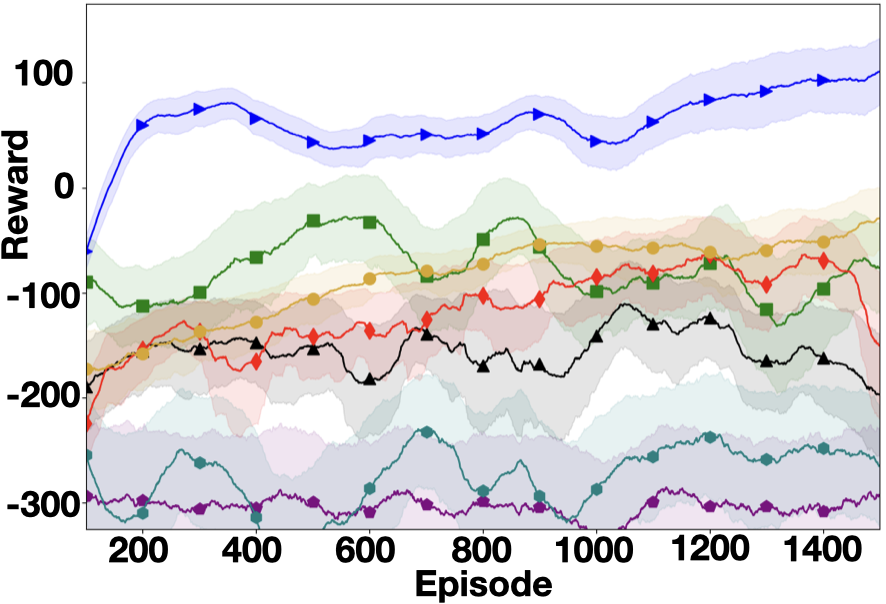}
        \caption{Lunar Lander}
        \label{fig:lunar-results}
    \end{subfigure}
    ~~
    \begin{subfigure}[b]{0.31\textwidth}
        \includegraphics[width=\textwidth]{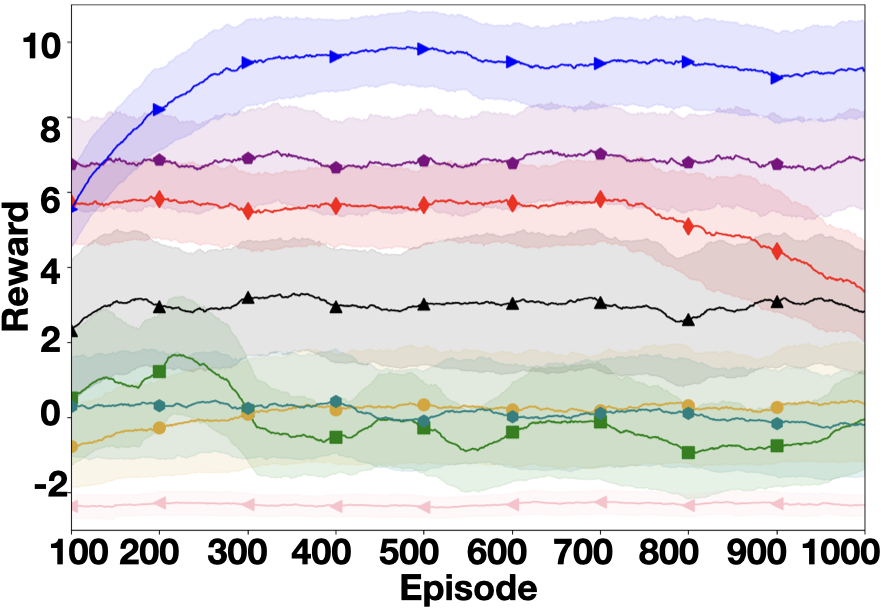}
        \caption{FindAndDefeatZerglings}
        \label{fig:micro-results}
    \end{subfigure}
\caption{A comparison of architectures on cart pole, lunar lander, and FindAndDefeatZerglings \cite{vinyals2017starcraft}. As the domain complexity increases, we see that intelligent initialization is increasingly important, and \textsc{ProLoNets} are the most effective method for leveraging domain expertise, and perform well even when domain expertise is unnecessary, as in cart pole.}
\label{fig:easy-results}
\end{figure*}

\subsection{Agent formulations}
\label{sub:bot-variants}

We compare several agents across our experimental domains. The first is a \textit{\textsc{ProLoNet}} agent as described above and with expert initialization. We also evaluate a multi-layer perceptron (\textit{MLP}) agent and a long short-term memory (\textit{LSTM})~\cite{hochreiter1997long} agent, both using ReLU activations~\cite{nair2010rectified}. We include comparisons to a \textsc{ProLoNet} with random initialization (\textit{Random \textsc{ProLoNet}}) as well as the \textit{Heuristic} used to initialize our agents. We compare to an IL agent trained with the \textit{LOKI} framework, in which the agent imitates for the first N episodes~\cite{cheng2018fast}, where N is a tuned hyperparameter, and then transitions to RL. The \textit{LOKI} agent supervises with the same heuristic that is used to initialize the \textit{\textsc{ProLoNet}} agent. Finally, although the original DJINN framework \cite{humbird2018deep} requires a decision tree learned over a labeled dataset, we extend the DJINN architecture to allow for initialization with a hand-crafted decision tree in order to compare to a \textit{DJINN} agent that is initialized using the same heuristic as \textit{LOKI} and \textit{\textsc{ProLoNet}}, but built with the DJINN architecture.

\subsection{Environments}
\label{sub:environment}
We consider four environments to empirically evaluate \textsc{ProLoNets}: cart pole, lunar lander, the FindAndDefeatZerglings minigame from the SC2LE~\cite{vinyals2017starcraft}, and a full game of SC2 against the in-game artificial intelligence (AI). These environments provide us with a steady increase in difficulty, from a toy problem to the challenging game of full SC2. These evaluations also showcase the ability of the \textsc{ProLoNet} framework to compete with state-of-the-art approaches in simple domains and excel in more complex domains. For the SC2 and SC2LE problems, we use the SC2 API\footnote{https://github.com/Blizzard/s2client-api} to manufacture 193D and 37D state spaces, respectively, and 44D and 10D action spaces, respectively. In the full SC2 domain, making the right parameter update is a significant challenge for RL agents. As such, we verify that the agent's parameter updates increase its probability of victory, and if a new update has decreased the agent's chances of success, then the update is rolled back, and the agent gathers experience for a different step, similar to the checkpointing approach in \citet{hosu2016playing}. 
%All experiments are run on an i7-7700K 4.2 GHz Quad-Core Processor, and we include code for our implementation and experiments.

\begin{table*}
\begin{center}
\begin{small}
\begin{sc}
% \begin{tabular}{lcccccc}
\begin{tabular}{*{7}{p{1.6cm}}}
 & & Random. & Shallow & & &\\
Domain & \textsc{ProLoNet} & \textsc{ProLoNet} & \textsc{ProLoNet} & N = 0.05 & N = 0.1 & N = 0.15 \\
\midrule
Cart Pole    & 449$\pm$15& 401$\pm$26& 415$\pm$ 27& 426$\pm$ 30& 369$\pm$ 28& 424$\pm$ 29 \\
Lunar & 86 $\pm$ 33&  55$\pm$19& 49$\pm$ 20& 50$\pm$ 22& 45$\pm$ 22& 45$\pm$ 22\\
Zerglings    & 8.9$\pm$1.5& -1.3$\pm$0.6& 8.8$\pm$1.5& 5.1$\pm$1.1& 5.9$\pm$1.2 & 4.1$\pm$1.1 \\
\end{tabular}
\end{sc}
\end{small}
\end{center}
\caption{\textsc{ProLoNet} ablation study of average cumulative reward. Units are in thousands.}
\label{tab:ablate}
\end{table*}
\paragraph{OpenAI Gym --}
\label{para:gym-results}

As depicted in Figure \ref{fig:cart-results} and \ref{fig:lunar-results}, \textsc{ProLoNets} are able to either match or exceed performance of standard reinforcement and imitation learning based RL architectures. Furthermore, we find that the \textsc{ProLoNet} architecture--even without intelligent-initialization--is competitive with baseline architectures in the OpenAI Gym. Running reward in these domains is averaged across five runs, as recommended by~\citet{henderson2018deep}. \textit{MLP} and \textit{LSTM} agents use 1-layer architectures which maintain input dimension until the output layer. We find success with intelligent initializations using as few as three nodes for the cart pole domain and as few as 10 nodes for the lunar lander. These results show that \textsc{ProLoNets} can leverage user knowledge to achieve superior results, and our ablation study results (Table \ref{tab:ablate}) show that the architecture is robust to sub-optimal initialization in these domains.

Even where intelligent initialization is not always necessary or where high-level instruction is difficult to provide, as in cart pole, it does not hinder RL from finding solutions to the problem. Further, while baselines appear unstable in these domains, potentially owing to missing implementation hacks and tricks \cite{engstrom2019implementation}, we observe that the \textsc{ProLoNet} approaches are able to succeed with the same PPO implementation and learning environment.

\paragraph{StarCraft II: FindAndDefeatZerglings --} For this problem, we assign an agent to each individual allied unit. The best-performing initialization in this domain has 6 decision nodes and 7 leaves. Running reward is depicted in Figure \ref{fig:micro-results}, again averaged over 5 runs. Intelligent initialization is crucial in this more complex domain, and the \textit{Random \textsc{ProLoNet}} fails to find much success despite having the same architecture as the \textit{\textsc{ProLoNet}}. \textit{LOKI} performs on par with the \textit{Heuristic} used to supervise actions, but \textit{LOKI} is unable to generalize beyond the \textit{Heuristic}. \textit{MLP} and \textit{LSTM} agents use a 7-layer architecture after a hyperparameter search, and we extend this to the full game of SC2. Importantly, this result (Figure \ref{fig:micro-results}) shows user-initialized \textsc{ProLoNets} can outperform our baselines and that this initialization is key to efficient exploration and learning. The importance of the initialization policy is again shown in Table \ref{tab:ablate}, where even negating 10\% of the agent's parameters results in a significantly lower average reward.

\paragraph{StarCraft II: Full Game --} After 5,000 episodes, no agent other than the \textit{\textsc{ProLoNet}} is able to win a single game against the in-game AI at the easiest setting. Even the \textit{LOKI} and \textit{DJINN} agents, which have access to the same heuristics used by the \textit{\textsc{ProLoNet}}, are unable to win one game. The \textit{\textsc{ProLoNet}}, on the other hand, is able to progress to the ``hard” in-game AI, achieving 100\% win rates against easier opponents as it progresses. Even against the ``hard” in-game AI, the \textit{\textsc{ProLoNet}} agent is able to double its win rate from initialization. This result demonstrates the importance of an intelligent initialization in complex domains, where only a very narrow and specific set of actions yield successful results. Access to oracle labeling (\textit{LOKI}) or a knowledge-based architecture (\textit{DJINN}) does not suffice; the agent requires the actual warm start of having intelligent rules built-in. Thus, we believe these results demonstrate that our novel formulation is singularly capable of harnessing domain knowledge.

\section{User study with non-experts}
\label{sec:user-study}

Our second evaluation investigates the utility of our framework with untrained humans providing the expert initialization for \textsc{ProLoNets}. As presented in Section \ref{subsec:study-results}, our user study shows that untrained users can leverage \textsc{ProLoNets} to train RL policies with superior performance. These results provide evidence that our approach can help democratize RL.

\begin{table}
\label{tab:macro-results}
\begin{center}
\begin{small}
\begin{sc}
\begin{tabular}{lccc}
&\textsc{ProLoNet}&\textsc{ProLoNet} at& All \\
AI Difficulty & (Ours) & Initialization & Others \\
\midrule
VeryEasy & \textbf{100\%}  & 14.1 \%& 0\%\\
Easy & \textbf{100\%}  & 10.9 \%& 0 \%\\
Medium & \textbf{82.2\%} & 11.3 \%& 0 \%\\
Hard & \textbf{26\%} & 10.7 \%& 0 \%\\
\end{tabular}
\end{sc}
\end{small}
\end{center}
\caption{Win rates against the StarCraft II in-game AI. ``All Others'' includes all agents in Section \ref{sub:bot-variants}.}
\end{table}

\paragraph{Hypotheses --}
\label{para:hypotheses}
We seek to investigate whether an untrained user can provide a useful initial policy for \textsc{ProLoNets}. Hypothesis 1 (\textbf{H1}): Expert initializations may be solicited from average users, requiring no particular training of the user, and these initializations are superior to random initializations. Hypothesis 2 (\textbf{H2}): RL can improve significantly upon these initializations, yielding superior policies after training.

\paragraph{Metrics --}
\label{para:metrics}
To test \textbf{H1}, we measure the reward over time for our best participant, all participants, and baseline methods. Testing \textbf{H2}, we measure the average reward for the first 50 and final 50 episodes for all agents specified by participants and our strongest baseline. Our metrics allow us to effectively examine our hypotheses in the context of expert initialization in our study domain.

\paragraph{Domain: Wildfire Tracking --}
\label{subsec:fire-sim}
We develop a Python simulator for a domain that is both suited to RL and of relevance to a wider audience: wildfire tracking. 
% A wildfire tracking problem is easier for a wider audience to understand and address, meaning we do not need to find a population of expert StarCraft II players or OpenAI Gym users. 
We randomly instantiate two fires and two drones in a 500x500 grid. The drones receive a 6D state as input, containing distances to fire centroids and Boolean flags for which a drone is the closest to each fire. The action space for drones is a 4D discrete decision of which cardinal direction to move into. Pre-made state checks include statements such as ``If I am the closest drone to Fire 2'' and ``If Fire 1 is to my west.'' The two drones are controlled by separate agents without communication, and network weights are shared. 
The reward function is the negative distance between drones and fire centroids, encouraging drones to follow the fire as closely as possible.

\subsection{Study details}
\label{subsec:study-details}
To solicit policy specifications from users, we designed a user interface that enabled participants to select from a set of pre-made state checks and actions. Participants were first briefed on the domain and shown a visualization and then asked to talk-through a strategy for monitoring the fires with two independent drones. After describing a solution and seeing the domain, participants were presented with the UI to build out their policies. As the participant selected options, those rules were composed into a decision tree. Once participants completed the study, we leveraged their policy specifications to initialize the structure and parameters of a \textsc{ProLoNet}. The \textsc{ProLoNet} was then deployed to the wildfire domain, where it further improved through RL. Our results are presented in Figure \ref{fig:firesim-results} and described below. We present both the highest performing participant (``Best''), as well as the median over all participants (``Median''), and compare against the agents presented in Section \ref{sub:bot-variants}. \textit{LOKI} and \textit{DJINN} agents use the ``Best'' participant policy specification as a heuristic.

\subsection{Study results}
\label{subsec:study-results}
Our IRB-approved study involved 15 participants (nine male, six female) between 21 and 29 years old ($M=24, SD=2$). The study took approximately 45 minutes, and participants were compensated for their time. Our pre-study survey revealed varying degrees of experience with robots and games, though we note that our participants were mostly computer science students. Importantly, we found that their prior experience with robots, learning from demonstration, or strategy games did not impact their ability to specify useful policies for our agents.
 
\begin{figure}[t]
\centering
        \includegraphics[width=0.7\linewidth]{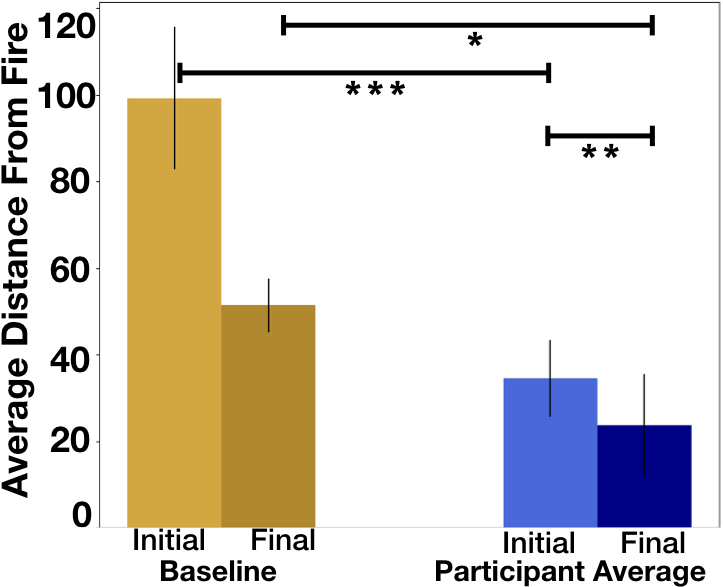}
        \caption{Initial and final distance between drones and wildfire centroids in our user study domain, where lower distance is better. Participant initializations are significantly better at tracking fires than random, showing that untrained users can leverage our approach to provide useful warm starts.} % . Lower is better.
        \label{fig:fire-bars}
    \end{figure}

\begin{figure}
    \centering
    \includegraphics[width=.93\linewidth]{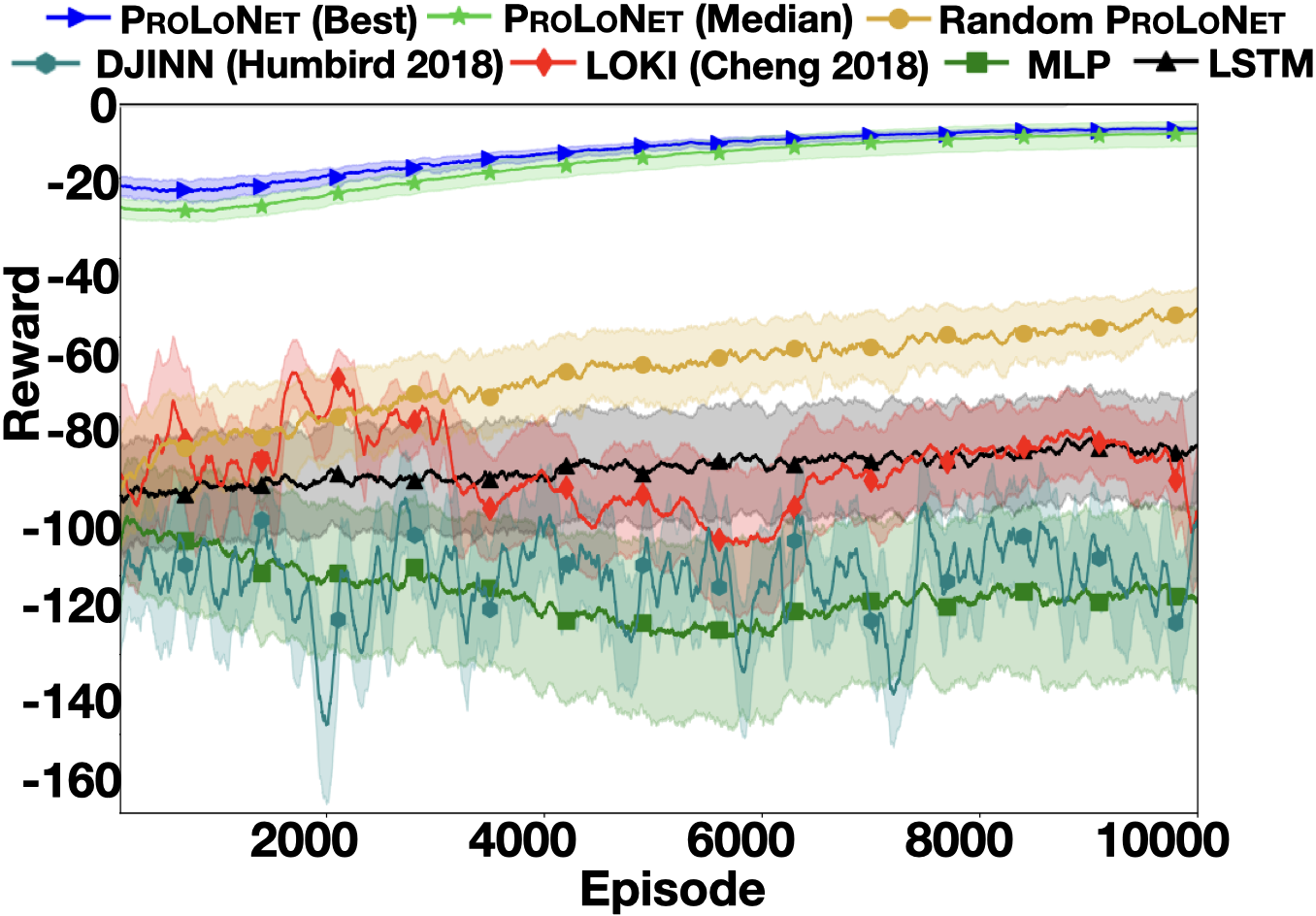}
    \caption{Wildfire tracking results, again demonstrating the importance of direct intelligent initialization (\textsc{ProLoNet}) rather than IL or random initialization.}
    \label{fig:firesim-results}
\end{figure}

Nearly all participants provided policy specifications that were superior to random exploration. After performing RL over participant specifications, we can see in Figure \ref{fig:firesim-results} that \textbf{intelligent initialization yields the most successful RL agents,} even from non-experts. We compare to the best performing baseline, \textit{Random \textsc{ProLoNet}} in Figure \ref{fig:fire-bars}. We can again see that the participants' initializations are not only better than random initialization, but are also better than the trained RL agent.
A Wilcoxon signed-rank test shows that our participants' initializations (Median = -23, IQR = 19) were significantly better than a baseline initialization  (Median = -87, IQR = 26), $W(15)=1.0, p < .001$. Our participants' agents (Median = -7.9 , IQR = 29) were also significantly better than a baseline (Median = -52, IQR = 7.9) after training, $W(15)=15.0, p = 0.011$. These results are significant after applying a Bonferroni correction to test the relative performance both before and after training. \textbf{This result supports hypothesis \textbf{H1}}, showing that average users can specify useful policies for RL agents to explore more efficiently than random search and significantly outperform baselines.

Furthermore, our participants' agents are significantly better post-training than at initialization, as shown by a Wilcoxon signed-rank test ($W(15)=4.0, p < 0.01$). \textbf{This finding supports hypothesis \textbf{H2}}, showing that RL improves on human specifications, not merely repeating what the humans have demonstrated. By combining human intuition and expertise with computation and optimization for long-term expected reward, we are able to produce agents that outperform both humans and traditional RL approaches.

Finally, we qualitatively demonstrate the utility of intelligent initialization and the \textsc{ProLoNet} architecture by deploying the top performing agents from each method to two drones with simulated fires to track. Videos of the top four agents are included as supplementary material.

\section{Discussion}
\label{sec:discussion}

We proposed two complementary evaluations of our proposed architecture, demonstrating the significance of our contribution. Through our first set of experiments on an array of RL benchmarks with a domain expert building heuristics, we empirically validated that \textsc{ProLoNets} are competitive with baseline methods when initialized randomly and, with a human initialization, outperforms state-of-the-art imitation and RL baselines. As we see in Figure \ref{fig:easy-results}, \textsc{ProLoNets} are as fast or faster than baseline methods to learn an optimal policy over the same environments and optimization frameworks. In our more complex domains, we identify the importance of an intelligent initialization. While the IL baseline performs well in the FindAndDefeatZerglings minigame, \textit{LOKI} cannot improve on the imitated policy. In the full game of SC2, no approach apart from our intelligently-initialized \textsc{ProLoNet} wins even a single game. The ability to leverage domain knowledge to initialize rules as well as structure, rather than simple architecture and routing information, as in DJINN, is a key difference that enables the success of our approach.

Through our user study, we demonstrated the practicality of our approach and shown that average participants, even those with no prior experience in the given domain, can produce policy specifications which significantly exceed random initialization ($p<0.05$). Furthermore, we have demonstrated that RL can significantly improve upon these policies, learning to refine ``good enough'' solutions into optimal ones for a given domain. This result shows us that our participants did not simply provide our agents with optimal solutions iterated upon needlessly. Instead, our participants provided good but sub-optimal starting points for optimization. These starting policies were then refined into a solution that was more robust than either the human's solution or the best baseline solution. Our study confirms that our approach can leverage readily available human initializations for success in deep RL, and moreover, that the combination of human initialization and RL yields the best of both worlds.

\section{Conclusion}

We present a new architecture for deep RL agents, \textsc{ProLoNets}, which permits intelligent initialization of agents. \textsc{ProLoNets} grant agents the ability to grow their network capacity as necessary, and are surprisingly capable even with random initialization. We show that \textsc{ProLoNets} permit initialization from average users and achieve a high-performing policy as a result of the blend of human instruction and RL. We demonstrate, first, that our approach is superior to imitation and reinforcement learning on traditional architectures and, second, that intelligent initialization allows deep RL agents to explore and learn in environments that are too complex for randomly initialized agents. Further, we have confirmed that we can solicit these useful warm starts from average participants and still develop policies superior to baseline approaches in the given domains, paving the way for reinforcement learning to become a more collaborative enterprise across a variety of complex domains.

\section*{Ethical Considerations}
Our work is a contribution targeted at democratizing reinforcement learning in complex domains. The current state of the art in reinforcement learning in complex domains requires compute time and power beyond the capacity of many labs, hand-engineering which is rarely explained publicly, or large labeled datasets which are not always shared. By providing a means for intelligent initialization by practitioners and improved exploration in many domains, we attempt to lower the barrier to entry for research in reinforcement learning and to broaden the number of potential applications of reinforcement learning to more grounded, real-world problems. While there are risks with any technology being misused, we believe the benefits of democratizing RL outweigh the risks. We posit that giving everyone the ability to use RL rather than just large corporations and select universities is a positive contribution to society.

\paragraph{Beneficiaries --} Our work seeks to improve and simplify reinforcement learning research for all labs and to take steps toward democratizing reinforcement learning for non-experts. We feel that the computational and dataset savings of our work stand to benefit all researchers within reinforcement learning.
\paragraph{Negatively affected parties --} We do not feel that any group of people or research direction is negatively impacted by this work. Our work is complementary to other explorations within reinforcement learning, and insights from imitation learning translate naturally into insights on the qualities of useful or harmful intelligent initializations.
\paragraph{Implications of failure --} While our method seeks to simplify reinforcement learning, in the worst case the initialization falls back to random and the learning agent is again faced with an intractable random exploration problem. Adversarial agents using our approach would be able to instantiate a worse-than-random agent, though our results imply that it is possible to overcome such an initialization in simple domains.
\paragraph{Bias and fairness --} Our work does rely on the ``bias'' of its initialization--that is, it is biased towards the actions which a human has pre-specified. While this biased exploration may fail to accurately explore or understand the intricacies of a complex domain, the alternative (years of compute with random exploration) is simply unavailable to many researchers. This bias may be overcome through diversification of intelligent initializations which may lead to a diversity of final strategies. However, the unification of such diverse policies into a single agent and the thorough study of diverse initializations is left to future work.
\bibliography{tree_nets}
\clearpage
\appendix
\section{\textsc{ProLoNet} Forward Pass}
An algorithmic step-through of the forward pass for the \textsc{ProLoNet} is provided in Algorithm \ref{alg:prolo-forward}. The example from the main paper is included here:
\begin{example}[\textsc{ProLoNet} Inference]
Consider an example cart pole state, X=[2, 1, 0, 3]. Following the equation in Line 3 of Algorithm \ref{alg:prolo-forward}, the network arrives at $\sigma([1,0,0,0]*[2,1,0,3]-0)=0.88$ for $D_0$, meaning “mostly true.” This decision probability propagates to the two leaf nodes using their respective paths (Lines 9-15 in Algorithm \ref{alg:prolo-forward}), making the output of the network a probability given by $(0.88 * [1, 0] + (1-0.88)*[0, 1]) = [0.88, 0.12]$. Accordingly, the agent selects the first action with probability 0.88 and the second action otherwise.
\end{example}

\begin{algorithm}[H]
  \caption{\textsc{ProLoNet} Forward Pass}
  \label{alg:prolo-forward}
\begin{algorithmic}
  \STATE {\bfseries Input:} Input Data $X$, \textsc{ProLoNet} $P$
  \FOR{$d_n \in D \in P$}
  \STATE $\sigma_n$ = $\sigma [\alpha(\vec{w_n}^T * \vec{X} - c_n)]$
  \ENDFOR
  \STATE $\vec{A}_{OUT} = $ Output Actions
  \FOR{$\vec{l_i} \in L$}
  \STATE Path to $\vec{l_i} = Z(L)$
  \STATE $z = 1$
  \FOR{$\sigma_i \in Z(L)$}
  \IF{$\sigma_i \,$ should be $\, TRUE \in Z(L)$}
  \STATE $z = z * \sigma_i$
  \ELSE
  \STATE $z = z * (1 - \sigma_i)$
  \ENDIF
  \ENDFOR
  \STATE $\vec{A}_{OUT} = \vec{A}_{OUT} + \vec{l_i} * z$
  \ENDFOR
  \STATE {\bfseries Return:} $\vec{A}_{OUT}$
\end{algorithmic}
\end{algorithm}

\section{Hyperparameters and Optimization Details}
All actors are updated with proximal policy optimization (PPO)~\cite{schulman2017proximal}. Notably, for the two SC2 domains, we find that multiplying the PPO update by the Kullback-Leibler divergence between old and new policies yields superior performance. The critic's loss function is the mean-squared error between the output of the critic and the reward from the state-action pair. All approaches are trained with RMSProp~\cite{tieleman2012lecture}. We set our reward discount factor to 0.99, learning rates to 1e-2 for Gym environments, and 1e-4 for the SC2 domains, following a hyperparameter search between 1-e2 and 1e-5. Update batch sizes dynamically grow as more replay experience is available. In all domains, the \textsc{ProLoNet} $\alpha$ parameter is initialized to 1.  Our agents utilize two separate networks: one for the actor and one for the critic. For our approach, the critic network is initialized as a copy of the actor as we do not solicit intelligent value predictions, only policies. Our dynamic growth hyperparameter $\epsilon$ is set to $\epsilon=0.1$ based upon experimental observation.

\section{Experimental Domain Details}
\subsection{Cart Pole}

Cart pole is an RL domain~\cite{barto1983neuronlike} where the object is to balance an inverted pendulum on a cart that moves left or right. The state space is a 4D vector representing \{\textit{cart position, cart velocity, pole angle, pole velocity}\}, and the action space is is \{\textit{left, right}\}. We use the cart pole domain from the OpenAI Gym~\cite{1606.01540}. 

For the cart pole domain, we set all agent's learning rates to 0.01, the batch size is set to dynamically grow as there is more replay experience available, we initialized $\alpha=1$, and each agent trains on all data gathered after each episode, then empties its replay buffer. All agents train on 2 simulations concurrently, pooling replay experience after each episode, and updating their policy parameters. For the $LOKI$ agent, we set $N$=200. All agents are updated according to the standard PPO loss function. We selected all parameters empirically to produce the best results for each method.

\subsection{Lunar Lander}
Lunar lander is the second domain we use from the OpenAI Gym~\cite{1606.01540}, and is based on the classic Atari game of the same name. Lunar lander is a game where the player attempts to land a small ship (the lander) safely on the ground, keeping the lander upright and touching down slowly. The 8D state consists of the lander's \{x, y\} position and velocity, the lander's angle and angular velocity, and two binary flags which are true when the left or right legs have touched down.

We use the discrete lunar lander domain, and so the 4D action space contains \{\textit{do nothing, left engine, main engine, right engine}\}. 
For the lunar lander domain, we set most hyperparameters to the same values as in the cart pole domain. The two exceptions are the number of concurrent processes, which we set to 4, and the $LOKI$ agent's $N$, which is set to 300. All agents use the standard PPO loss function. 
\subsection{FindAndDefeatZerglings}

FindAndDefeatZerglings is a minigame from the SC2LE designed to challenge RL agents to learn how to effectively micromanage their individual attacking units in SC2. The agent controls three attacking units on a small, partially-observable map, and must explore the map while killing enemy units. The agent receives +1 reward for each enemy unit that is killed, and -1 for each allied unit that is killed. Enemy units respawn in random locations, and so the best agents are ones that continuously explore and kill enemy units until the three minute timer has elapsed.

We leverage the SC2 API \footnote{https://github.com/Blizzard/s2client-api} to manufacture a 37D state which contains \{\textit{x\_position, y\_position, health, weapon\_cooldown}\} for three allied units, and \{\textit{x\_position, y\_position, health, weapon\_cooldown, is\_baneling}\} the five nearest visible enemy units. Missing information is filled with -1. Our action space is 10D, containing move commands for north, east, south, west, attack commands for each of the five nearest visible enemies, and a ``do nothing'' command. For this problem, we assign an agent to each individual allied unit, which generates actions for only that unit. Experience from each agent stops accumulating when the unit dies. All experience is pooled for policy updates after each episode, and parameters are shared between agents.

For the SC2LE minigame, we set all agents' learning rates to 0.001, we again initialized $\alpha=1$, and the batch size to 4. Each agent trains on replay data for 50 update iterations per episode, and pools experience from 2 concurrent processes. The $LOKI$ agent's $N$, is set to 500. The agents in this domain update according to the loss function in Equation \ref{eqn:KLLoss}.

\begin{equation}
\label{eqn:KLLoss}
L(a, s, \pi_{new}, \pi_{old})  = \frac{(A*log(a|\pi_{new}))}{KL(P(\vec{a}|\pi_{new}, s), P(\vec{a}|\pi_{old}, s))}
\end{equation}

Where $A$ is the advantage gained by taking action $a$ in state $s$, $\pi_{new}$ is the current set of model parameters, and $\pi_{old}$ is the set of model parameters used during the episode which generated this state-action pair. $\vec{a}$ is the probability distribution over all actions that a policy $\pi$ yields given state $s$. As in prior work, the advantage $A$ is calculated by subtracting the reward (obtained by taking action $a$ in state $s$) from the value prediction for taking action $a$ in state $s$, given by a critic network.
\subsection{SC2 Full Game}
Our simplified StarCraft 2 state contains:
\begin{itemize}
    \item \textit{Allied Unit Counts}: A 36x1 vector in which each index corresponds to a type of allied unit, and the value corresponds to how many of those units exist.

    \item \textit{Pending Unit Counts}: As above, but for units that are currently in production and do not exist yet.

    \item \textit{Enemy Unit Counts}: A 112x1 vector in which each index corresponds to a type of unit, and the value corresponds to how many of those types are visible.

    \item \textit{Player State}: A 9x1 vector of specific player state information, including minerals, vespene gas, supply, etc. 
\end{itemize}

The disparity between allied unit counts and enemy unit counts is due to the fact that we only play as the Protoss race, but we can play against any of the three races.

The number of actions in SC2 can be well into the thousands if one considers every individual unit's abilities. As we seek to encode a high-level strategy, rather than rules for moving every individual unit, we restrict the action space for our agent. Rather than using exact mouse and camera commands for individual units, we abstract actions out to simply: ``Build Pylon.'' As such, our agents have 44 available actions, including 35 building and unit production commands, 4 research commands, and 5 commands for attack, defend, harvest resources, scout, and do nothing.

For the full SC2 game, we set all agents' learning rates to 0.0001, we again initialized $\alpha=1$, set the batch size to 4, and updates per episode to 8. We run 4 episodes between updates, and set the $LOKI$ $N$=1000. Agents train for as long as necessary to achieve a $>80\%$ win-rate against the easiest AI, then move up to successive levels of difficulty as they achieve $>80\%$ win-rates. The agents in this domain update according to the loss function in Equation \ref{eqn:KLLoss}.

\subsection{User Study Domain: Wildfire Tracking}
The objective in the wildfire tracking domain is to keep two drones on top of two fire centroids as they progress through the map. The task is complicated by the fact that the two drones do not communicate, and do not have complete access to the state of the world. Instead, they have access to a 6D vector containing \{ $D_N(F_1)$, $D_W(F_1)$, $D_N(F_2)$, $D_W(F_1)$, $C(F_1)$, $C(F_2)$ \} where $D_N$ is the ``distance to the north`` function and $C(F_1)$ is the ``closer to fire 1'' boolean flag.

The actions available to the drones include move commands in four directions: north, east, south, and west.

\section{Initialization Heuristics in Experimental Evaluation}
\subsection{Cart Pole Heuristics}
\label{sec:cart-heuristics}
\begin{figure*}[b]
    \centering
    \includegraphics[width=\textwidth]{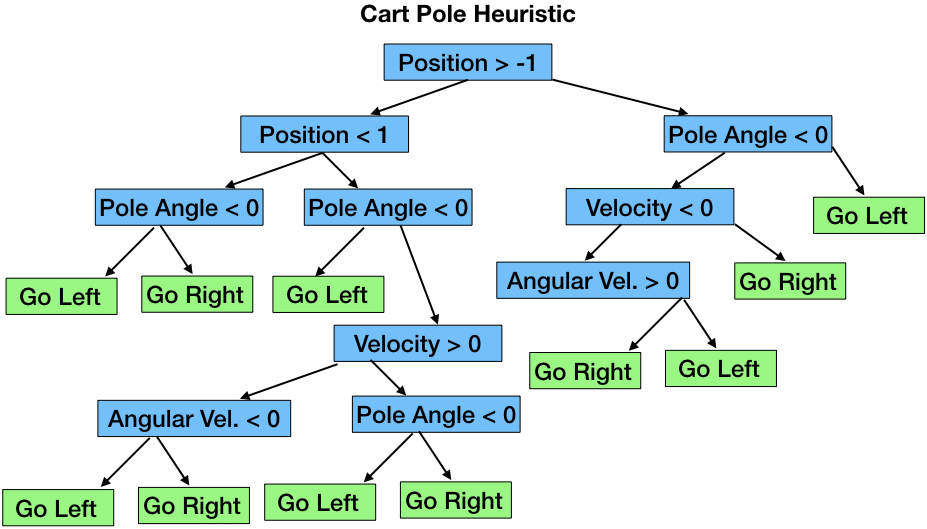}
    \caption{Visualization of the heuristics used to initialize the cart pole \textsc{ProLoNet}, and to train the LOKI agent}
    \label{fig:cart-heuristic}
\end{figure*}

We use a simple set of heuristics for the cart pole problem, visualized in Figure \ref{fig:cart-heuristic}. If the cart is close enough to the center, we move in the direction opposite to the lean of the pole, as long as that motion will not push us too far from the center. If the cart is close to an edge, the agent attempts to account for the cart's velocity and recenter the cart, though this is often an unrecoverable situation for the heuristic. The longest run we saw for a \textsc{ProLoNet} with no training was about 80 timesteps.

\subsection{Lunar Lander Heuristics}
\label{sec:lunar-heuristics}
\begin{figure*}
    \centering
    \includegraphics[width=\textwidth]{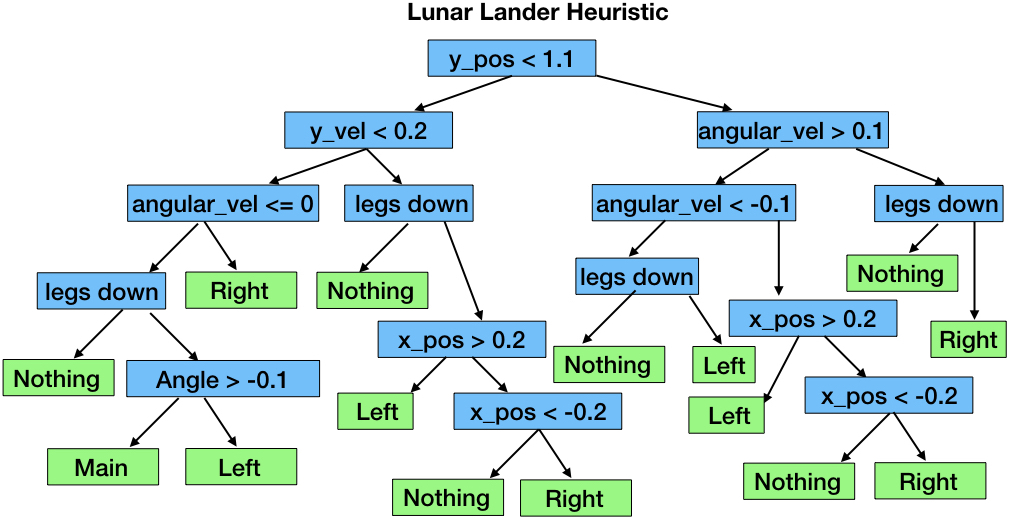}
    \caption{Visualization of the heuristics used to initialize the lunar lander \textsc{ProLoNet}, and to train the LOKI agent}
    \label{fig:lunar-heuristic}
\end{figure*}

For the lunar lander problem, the heuristic rules are split into two primary phases. The first phase is engaged at the beginning of an episode while the lander is still high above the surface. In this phase, the lander focuses on keeping the lander's angle as close to 0 as possible.
Phase two occurs when the lander gets closer to the surface, and the agent then focuses on keeping the y\_velocity lower than 0.2. As is depicted in Figure \ref{fig:lunar-heuristic}, there are many checks for both lander legs being down. We found that both $LOKI$ and \textsc{ProLoNets} were prone to landing successfully, but continuing to fire their left or right boosters. In an attempt to ameliorate this problem, we added the extra ``legs down'' checks.

\subsection{FindAndDefeatZerglings Heuristics}
\label{sec:micro-heuristics}
\begin{figure*}
    \centering
    \includegraphics[width=\textwidth]{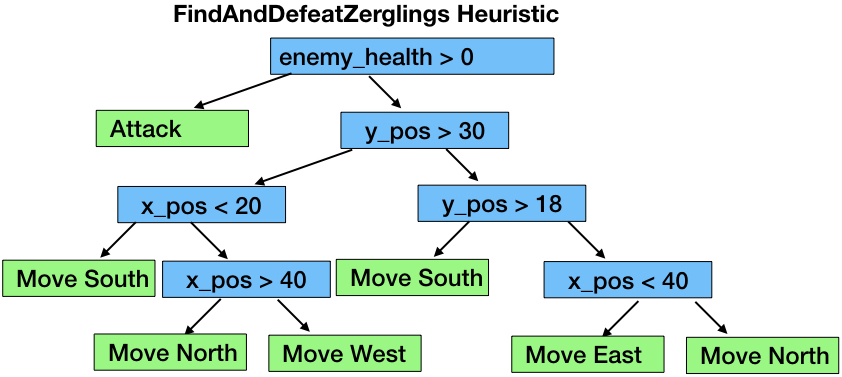}
    \caption{Visualization of the heuristics used to initialize the FindAndDefeatZerglings \textsc{ProLoNet}, and to train the LOKI agent}
    \label{fig:micro-heuristic}
\end{figure*}

For the SC2LE minigame, the overall strategy of our heuristic is to stay grouped up and fight or explore as a group. As such, the first four checks are all in place to ensure that the marines are all close to each other. After they pass the proximity checks, they attack whatever is nearest. If nothing is nearby, they will move in a counter-clockwise sweep around the periphery of the map, searching for more zerglings. Our heuristic is shown in Figure \ref{fig:micro-heuristic}.

\subsection{SC2 Full Game Heuristics}
\label{sec:macro-heuristics}
\begin{figure*}
    \centering
    \includegraphics[width=\textwidth]{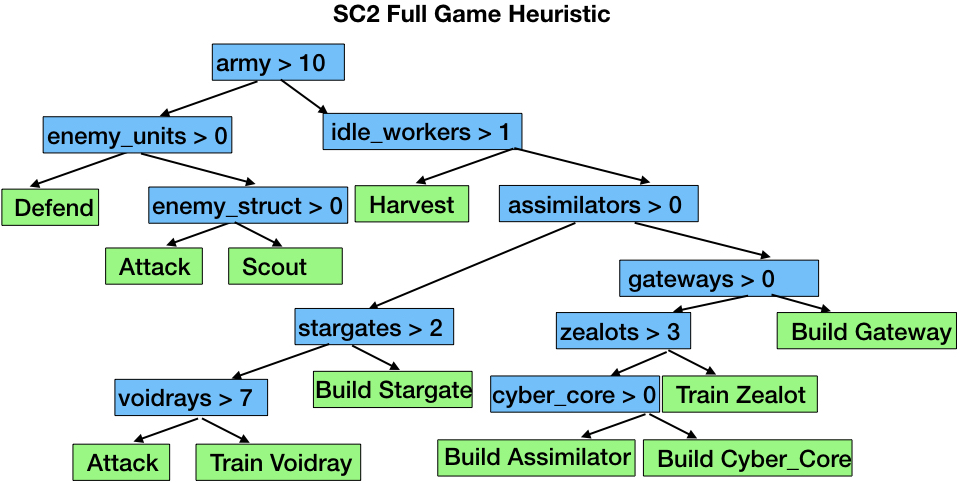}
    \caption{Visualization of the heuristics used to initialize the SC2 full game \textsc{ProLoNet}, and to train the LOKI agent}
    \label{fig:macro-heuristic}
\end{figure*}

The SC2 full game heuristic first checks for important actions that should always be high priority, such as attacking, defending, harvesting resources, and scouting. Once initial checks for these are all passed, the heuristic descends into the build order, where it simply uses building or unit count checks to determine when certain units should be built or trained. After enough attacking units are trained, the heuristic indicates that it is time to attack. The SC2 full game heuristic is depicted in Figure. \ref{fig:macro-heuristic}.

\section{Architectures for Algorithms in Experimental Evaluation}
\label{sec:fc-lstm-arches}
In this section we briefly overview the $MLP$ and $LSTM$ action network information. The $LOKI$ agent maintained the same architecture as the $MLP$ agent.

\subsection{Cart Pole}
The cart pole $MLP$ network is a 3-layer network following the sequence:

4x4 -- 4x4 -- 4x2.

We experimented with sizes ranging from 4-64 and numbers of hidden layers from 1 to 10, and found that the small network performed the best.

The $LSTM$ network for cart pole is the same as the $MLP$ network, though with an LSTM unit inserted between the first and second layers. The LSTM unit's hidden size is 4, so the final sequence is:

4x4 -- LSTM(4x4) -- 4x4 -- 4x2.

We experimented with hidden-sizes for the LSTM unit from 4 to 64, though none were overwhelmingly successful, and we varied the number of layers after the LSTM unit from 1-10.

The \textsc{ProLoNet} agent for this task used 9 decision nodes and 11 leaves. For the deepening experiment, we tested an agent with only a single node and 2 leaves, and found that it still solved the task very quickly. We tested randomly initialized architectures from 1 to 9 nodes and from 2 to 11 leaves, and we found that all combinations successfully solved the task.

\subsection{Lunar Lander}
The lunar lander $MLP$ network is a 3-layer network, following the sequence:

8x8 -- 8x8 -- 8x4. 

We again experimented with sizes from 8-64 and number of hidden layers from 2 to 11.

The $LSTM$ network for lunar lander mimics the architecture from cart pole. The LSTM unit's hidden size is 8, so the final sequence is:

8x8 -- LSTM(8x8) -- 8x4.

We experimented with hidden-sizes for the LSTM unit from 8 to 64, and again we varied the number of layers succeeding the LSTM unit from 1 to 10.

The \textsc{ProLoNet} agent for this task featured 14 decision nodes and 15 leaves. We experimented with intelligent initialization architectures ranging from 10 nodes to 14 and from 10 to 15 leaves, and found little difference between their performances. The additional nodes were an attempt to encourage the agent to ``do nothing'' once successfully landing, as the agent had a tendency to continue shuffling left-right after successfully touching down. 

\subsection{FindAndDefeatZerglings}
We failed to find a $MLP$ architecture that succeeded in this task, and so we choose one that compromised between the depth of the \textsc{ProLoNet} and the simplicity that $MLP$ agents seemed to prefer for toy domains. The final network is a 7-layer network with the following sequence:

37x37 -- 37x37 -- 37x37 -- 37x37 -- 37x37 -- 37x37 -- 37x10.

We choose to keep the size to 37 after testing 37 and 64 as sizes, and deciding that trying to get as close to the \textsc{ProLoNet} architecture was the best bet.

The $LSTM$ network for FindAndDefeatZerglings features more hidden layers than the $LSTM$ for lunar lander and cart pole. The hidden size is set to 37, and the LSTM unit is followed by 5 layers. The final sequence is:

37x37 -- LSTM(37x37) -- 37x37 -- 37x37 -- 37x37 -- 37x37 -- 37x10.

We experimented with hidden-sizes for the LSTM unit from 37 to 64 and varied the number of successive layers from 4-10.

The \textsc{ProLoNet} agent for FindAndDefeatZerglings featured 10 nodes and 11 leaves. We tested architectures from 6 to 15 nodes and from 7 to 13 leaves, and found that the initialized policy and architecture had more of an immediate impact for this task. The 7 node policy allowed agents to spread out too much, and they died quickly, whereas the 15 node policy had agents moving more than shooting, and they would walk around while being overrun. 
\subsection{SC2 Full Game}
We again failed to find a $MLP$ architecture that succeeded in this task, and so used a similar architecture to that of the FindAndDefeatZerglings task. The 7-layer network is of the sequence:

193x193 -- 193x193 -- 193x193 -- 193x193 -- 193x193 -- 193x193 -- 193x44.

We again experimented with a variety of shapes and number of layers, though none succeeded.

Again, the $LSTM$ network shadows the $MLP$ network for this task. As in the FindAndDefeatZerglings task, we experimented with a variety of LSTM hidden unit sizes, hidden layer sizes, and hidden layer numbers. The final architecture reflects the FindAndDefeatZerglings sequence:

193x193 -- LSTM(193x193) -- 193x193 -- 193x193 -- 193x193 -- 193x193 -- 193x44.

The \textsc{ProLoNet} agent for the SC2 full game featured 10 nodes and 11 leaves. We tested architectures from 10 to 16 nodes and from 1 to 17 leaves, and found that the initialized policy and architecture was not as important for this task as it was for the FindAndDefeatZerglings task. As long as we included a basic build order and the ``attack'' command, the agent would manage to defeat the VeryEasy in-game AI at least 10\% of the time. We found that constraining the policy to fewer nodes and leaves provided less noise as updates progressed, and kept the policy close to initialization while also providing improvements. An initialization with too many parameters often seemed to degrade quickly, presumably due to small changes over many parameters having a larger impact than small changes over few parameters.

\subsection{User Study Domain: Wildfire Tracking}
The wildfire tracking domain has a similar state-action space to the lunar lander domain. Therefore, we reuse architecture specifics from the lunar lander architecture sweep. The $MLP$ agent's action network is a 3-layer network, following the sequence:

6x6 -- 6x6 -- 6x4. 

The $LSTM$ network for the wildfire tracking problem also mimics the architecture from the $LSTM$ agent on the lunar lander problem. The LSTM unit's hidden size is 6, so the final sequence is:

6x6 -- LSTM(6x6) -- 6x4.

\textsc{ProLoNet} initializations varied substantially on this domain, though most produced initializations which eventually became nearly-optimal policies.

\section{Policy Divergence After Training}
\begin{figure*}[h]
\centering
\begin{subfigure}[b]{\textwidth}
    \centering
    \includegraphics[width=0.5\textwidth]{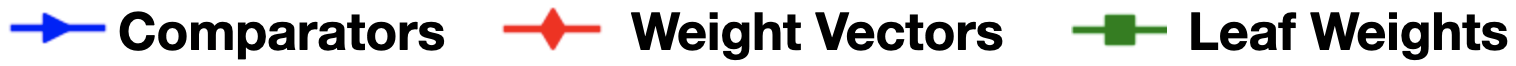}
\end{subfigure}
    \begin{subfigure}[b]{0.31\textwidth}
        \includegraphics[width=\textwidth]{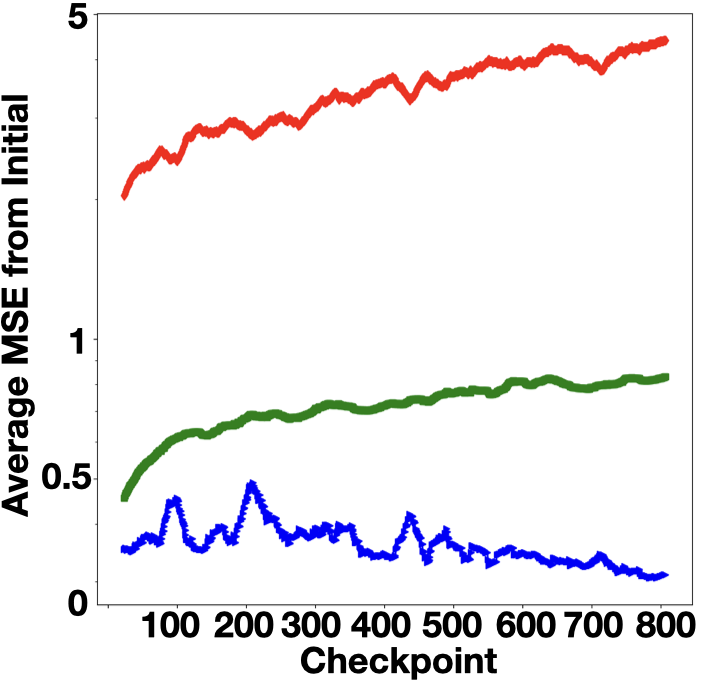}
        \caption{Cart Pole}
        \label{fig:cart-divergence}
    \end{subfigure}
    ~~
    \begin{subfigure}[b]{0.31\textwidth}
        \includegraphics[width=\textwidth]{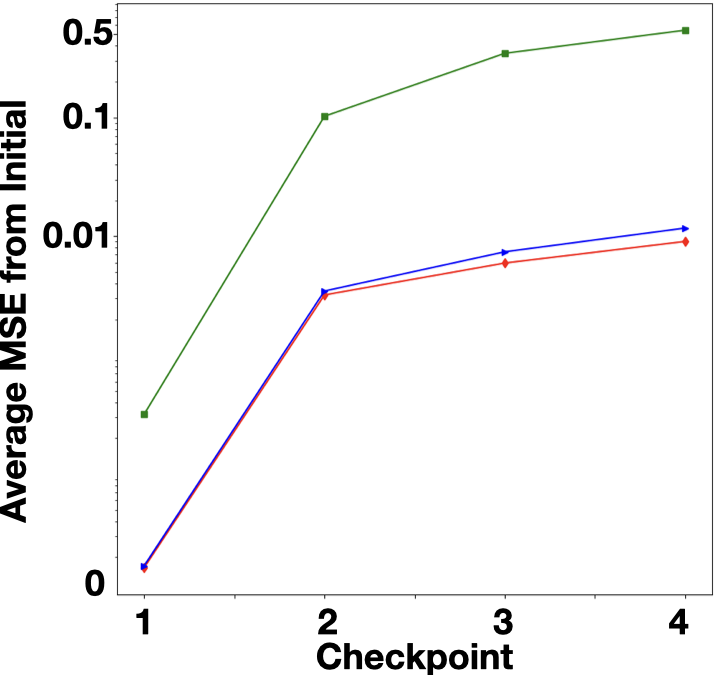}
        \caption{Lunar Lander}
        \label{fig:lunar-divergence}
    \end{subfigure}
    ~~
    \begin{subfigure}[b]{0.31\textwidth}
        \includegraphics[width=\textwidth]{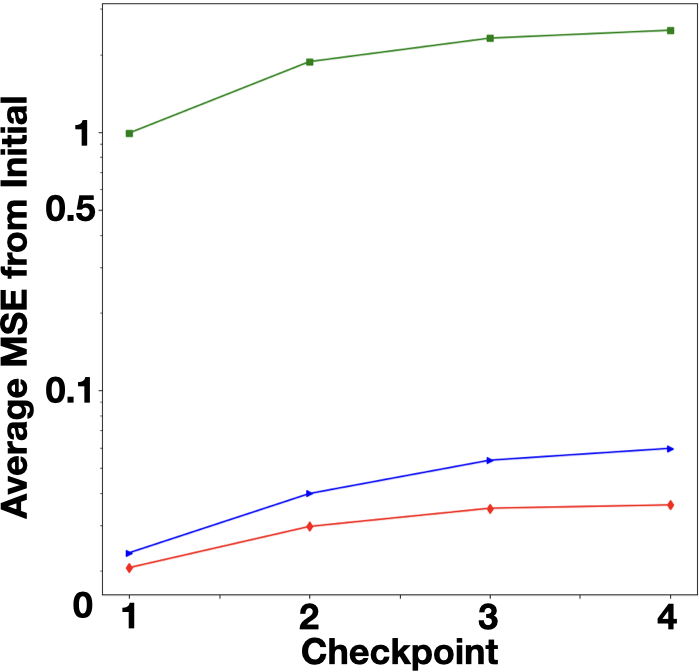}
        \caption{FindAndDefeatZerglings}
        \label{fig:micro-divergence}
    \end{subfigure}
\caption{A comparison of divergence of policies from initialization on cart pole, lunar lander, and FindAndDefeatZerglings.}
\label{fig:divergence-results}
\end{figure*}

We observe that finished policies are not rehashes of the originals; rather, they change and deviate from the original throughout training. In the figures below, we compare checkpointed models to the original initialization. The x-axis corresponds to how far along in the experiment the checkpoint is, and the y-axis corresponds to the average mean-squared error between the initialization and the checkpoint. As the mean-squared error for the weight vectors, comparator values, and leaf weights can be markedly different, we use a logarithmic scale on the y-axis so that the trends can clearly be seen regardless of the raw value. Results are shown in Figure \ref{fig:divergence-results}. Note that we checkpoint after every 25\% of training or when the agent has ``solved'' the domain by the OpenAI Gym standards (500 for cart pole, 200+ for lunar lander), accounting for the greater density of checkpoints in the cart pole domain.

\section{User Study Examples}
Our user study includes soliciting natural language instructions from participants as a way to prime them to think through policies before interacting with our interface. We include excerpts below to demonstrate the variety of participants in our study.

\subsection{Before Using the User Interface}
\begin{flushenumerate}
    \item ``If it is the closet drone to the first fire, go to the first fire. 
Else, if it is the closet drone to the second fire, go to the second fire. 
Else, (it is not closet drone to the first fire and second fire), go to the second fire.''
    \item `` If you are the closest drone to the first fire:
	Determine how far north, south, east, and west you are from the fire
	Pick a direction to travel first (north/south, east/west) and see if the distance in that axis between you and the fire gets smaller. 
If so, continue that direction but add some angle towards a direction in another axis (e.g. if you determined that traveling north is getting you closer to the fire, pick northwest or northeast to start driving towards). If the new axis addition is not getting you closer to the fire, then decrease your addition of that angle. If you reach zero, then switch to the other axis (i.e. if going northwest does not help, switch to northeast).
If your original direction (pure north, south, east, or west) does not get you closer to the fire, go the opposite direction in that axis and see if the distance in that axis decreases. 

If you are the closest drone to the second fire:
	Run same instructions as for tracking the first fire.

If you are the closest drone to both fires:
	Pick the fire that’s the closest to you and run the same instructions as you would for tracking first fire.

If you’re not the closest drone to either fire:
	Move randomly until you are the closest drone to one of the fires''
\end{flushenumerate}

\subsection{While Using the User Interface}
\begin{flushenumerate}
    \item ``If closest to fire 1, if fire 1 is south, move south. If fire 1 is north, move north. If fire 1 is east, move east. If fire 1 is west, move west. Otherwise, move east. If not closest to fire 1, if fire 2 is south, move south. If fire 2 is north, move north. If fire 2 is east, move east, otherwise just move west.''
    \item ``So let’s start with fire 1. So if I’m the closest drone to fire 1, if fire 1 is to my south, move south. Else if fire 1 is to my north, move north. Else if fire 1 is to my east, move east. Else if fire 1 is to my west, move west. If I’m not the closest drone to fire 1, check if I’m the closest drone to fire 2. If fire 2 is to your south, move south. If fire 2 is to your north, move north. If fire 2 is to your east, move east. If fire 2 is to your west, move west. If not, just take a random action. If you’re not closest to fire 1 or fire 2, move north.''
\end{flushenumerate}

\subsection{After Using the User Interface}
\begin{flushenumerate}
    \item ``If you are the closest to fire 1, check if the fire1 is north of you. If so, move north. If not, check if the drone is south of you. If so, move south. If not, check if the drone is east of you. If so, move east. If not, check if the drone is west of you. If so move east, otherwise do a random action.
If you are NOT the closest to fire 1, check if fire 2 is north of you. If so, move north. If not, check if the drone is south of you. If so, move south. If not, check if the drone is east of you. If so, move east. If not, check if the drone is west of you. If so move east, otherwise do a random action.''
    \item `` If it is the closet drone to the first fire, 
If the first fire’s north direction is positive, move north. 
If the first fire’s north direction is negative, move south. 
If the first fire’s east direction is positive, move east. 
If the first fire’s east direction is negative, move west. 
Else, if it is the closet drone to the second fire, go to the second fire. 
If the second fire’s north direction is positive, move north. 
If the second fire’s north direction is negative, move south. 
If the second fire’s east direction is positive, move east. 
If the second fire’s east direction is negative, move west. 
Else, (it is not closet drone to the first fire and second fire), go to the second fire. 
	If the first fire’s north direction is positive, move north. 
If the first fire’s north direction is negative, move south. 
If the first fire’s east direction is positive, move east. 
If the first fire’s east direction is negative, move west. ''
\end{flushenumerate}
\end{document}